\documentclass{article}
\usepackage{iclr2025_conference,times}

\usepackage{amsmath,amsfonts,bm}

\def\eqref#1{equation~\ref{#1}}
\def\1{\bm{1}}

\DeclareMathAlphabet{\mathsfit}{\encodingdefault}{\sfdefault}{m}{sl}
\SetMathAlphabet{\mathsfit}{bold}{\encodingdefault}{\sfdefault}{bx}{n}

\DeclareMathOperator*{\argmin}{arg\,min}

\usepackage{hyperref}
\usepackage{url}
\usepackage{graphicx}
\usepackage{cleveref}
\usepackage{subcaption}
\usepackage{booktabs}

\title{Adapting a World Model for\\Trajectory Following in a 3D Game}

\author{%
  Marko Tot$^{1,2}$\thanks{Equal contribution}\hspace{4pt}\thanks{This work was conducted while at Microsoft Research}\\
  \And
  Shu Ishida$^{1,3}$\footnotemark[1]\hspace{4pt}\footnotemark[2]\\
  \And
  Abdelhak Lemkhenter$^1$\\
  \AND
  David Bignell$^1$\\
  \And
  Pallavi Choudhury$^1$\\
  \And
  Chris Lovett$^1$\\
  \AND
  Luis França$^1$\\
  \And
  Matheus R. F. Mendonça$^1$\\
  \And
  Tarun Gupta\footnotemark[2]\\
  \And
  Darren Gehring$^1$\\
  \AND
  Sam Devlin$^1$\\
  \And
  Sergio Valcarcel Macua$^1$\\
  \And
  Raluca Stevenson$^1$\\
  \AND
  $^1$Microsoft Research \\
  \And
  $^2$Queen Mary University of London \\
  \And
  $^3$University of Oxford \\
}

\iclrfinalcopy % Uncomment for camera-ready version, but NOT for submission.
\begin{document}

\maketitle

\begin{abstract}
Imitation learning is a powerful tool for training agents by leveraging expert knowledge, and being able to replicate a given trajectory is an integral part of it. In complex environments, like modern 3D video games, distribution shift and stochasticity necessitate robust approaches beyond simple action replay. In this study, we apply Inverse Dynamics Models (IDM) with different encoders and policy heads to trajectory following in a modern 3D video game -- Bleeding Edge. 
Additionally, we investigate several future alignment strategies that address the distribution shift caused by the aleatoric uncertainty and imperfections of the agent. We measure both the trajectory deviation distance and the first significant deviation point between the reference and the agent's trajectory and show that the optimal configuration depends on the chosen setting. Our results show that in a diverse data setting, a GPT-style policy head with an encoder trained from scratch performs the best, DINOv2 encoder with the GPT-style policy head gives the best results in the low data regime, and both GPT-style and MLP-style policy heads had comparable results when pre-trained on a diverse setting and fine-tuned for a specific behaviour setting. 

\end{abstract}

\section{Introduction} \label{sec:introduction}

Using video games as a testbed for game-playing agents has been a thoroughly studied area. Although imitation learning and reinforcement learning have been applied, most of these algorithms~\citep{alphastar, openai_five, article} focused on superhuman behaviour, rather than matching human play style. 
Research on human-like play primarily leverages imitation learning, where the most popular techniques revolve around learning from demonstration~\citep{10.1145/1015330.1015430, DBLP:journals/corr/HoE16} and learning from observations~\citep{DBLP:journals/corr/abs-1807-06158,yang2019imitation}.

In this work, we use learning from demonstrations to replicate a recorded trajectory in a complex 3D video game. In simple environments, trajectory replication can often be achieved by directly replaying recorded actions. However, in stochastic settings, naive action playback fails, as small variations in state transitions can lead to significant deviations from the intended trajectory. 

To address this challenge, we adapt a pre-trained World Model to construct an Inverse Dynamics Model (IDM)~\citep{lamb2023guaranteed}. We evaluate the effectiveness of world model embeddings compared to alternative encoders and explore future alignment strategies to improve the replication of recorded trajectories in the video game Bleeding Edge.
Our evaluation involves two types of policy heads -- an autoregressive transformer~\citep{radford_language_2019} and a feed-forward network -- paired with three encoder types: a pre-trained game-specific world model, a general pre-trained encoder, and a ConvNeXt trained from scratch. This results in six distinct model configurations.

We evaluate these six model variants across three experimental settings: 1) General - trained on a large corpus of general gameplay trajectories, and evaluated on held-out trajectories, 2) Specific - trained on a small set of similar trajectories that exhibit the same behaviour, and evaluated on the same class, and 3) Fine-tuned - where we pre-train the model using 1) and then fine-tune and evaluate it using 2).

In summary, our contributions are as follows:
\begin{itemize}
    \item Adapting a pre-trained world model for imitation learning in a downstream task of trajectory following. 
    \item Conducting an empirical analysis of different model configurations across General, Specific and Fine-tuned settings in a complex 3D video game.
    \item Investigating the impact of design choices, such as using single observation vs sequence of observations as the model inputs, and the inclusion action inputs.
    \item Exploring different future conditioning strategies to mitigate distribution shifts and improve long-term trajectory alignment.
\end{itemize}

\section{Related Work} \label{sec:related_work}

\paragraph{World Models}
World Models~\citep{NEURIPS2018_2de5d166} have shown strong capabilities in simulating environments. They have been successfully applied to games such as Doom~\citep{valevski2024diffusionmodelsrealtimegame}, Atari~\citep{micheli2023transformers} and Counter Strike~\citep{alonso2024diffusion}, allowing users to interact with the game without reliance on an underlying game engine. Recent studies have shown that world models follow similar scaling laws as Large Language Models~\citep{pearce2024scalinglawspretrainingagents}, exhibit zero-shot generalisation to unseen tasks~\citep{xu2022learning}, and can scale to multi-game environments~\citep{pmlr-v235-bruce24a}.

\paragraph{Imitation Learning}
Imitation learning enables agents to learn tasks by observing expert demonstrations rather than relying on reward signals or direct exploration, as in reinforcement learning. The goal is to replicate observed behaviours by learning policies that map states to actions. One of the most widely used approaches is Behavioural Cloning (BC)~\citep{Torabi2018BehavioralCF}, which trains models on offline datasets to mimic human behaviour. BC has been applied in domains such as autonomous driving~\citep{6796843}, robotics~\citep{pmlr-v164-florence22a} as well as game-playing agents in video games~\citep{DBLP:journals/corr/abs-2004-00981, Vinyals2019GrandmasterLI}. 
While much of the research in imitation learning for games focuses on optimising policies for high performance~\citep{Vinyals2019GrandmasterLI, DBLP:journals/corr/abs-1011-0686, DBLP:journals/corr/HoE16, 9893617}, some works have explored capturing diverse played behaviours and playstyles.
\citet{6ad216893c9947bbb746862e109dccf0} investigated the use of Dynamic Time Warping~\citep{Müller2007} to imitate a given playstyle, while \citet{pearce2023imitating} used Diffusion Models to capture multi-modal behavioural patterns. These studies highlight the potential of imitation learning to generate human-like gameplay beyond optimal strategies.

\paragraph{Inverse Dynamics Models}

Inverse Dynamics Model (IDM) is commonly used to condition agents by predicting the action required to transition from one state to another. \citet{paster2021planning} applied an IDM to task condition the agent in a visual domain to predict a sequence of actions.  \citet{yang2019imitation} used an IDM, as a tool, to minimise the disagreement between the expert and the agent during training.  \citet{Pavse2019RIDMRI} combined an IDM with reinforcement learning to match the expert's trajectory. They use an inverse dynamics model to infer the action that should be taken to traverse from the learner's to the expert's trajectory in robotic control domains. 
While our approach shares similarities with \citet{Pavse2019RIDMRI}, there are key distinctions. Our method focuses on directly matching reference trajectories, whereas their objective was to generate high-scoring trajectories when conditioned on the expert behaviour. Additionally, we train our IDM on offline datasets, avoiding the need for an RL feedback loop to avoid the time cost of querying the environment.

\section{Method} \label{sec:method}
A conventional IDM encodes the current observation and next observation to predict the action that has been taken. IDM-$K$ generalises this approach, by shifting the future conditioning $K$ steps ahead.
Rather than relying on a single observation pair, we extend the notion of IDM-$K$ by conditioning on past and future trajectory sequences - comprising of both observations and actions - to improve the temporal consistency and long-term dependency. This is particularly important in partially observable environments, such as video games, where single-frame conditioning can be insufficient for accurate decision-making.

\subsection{Models}

\begin{figure} [!t]
     \centering
     \includegraphics[width=0.78\textwidth]{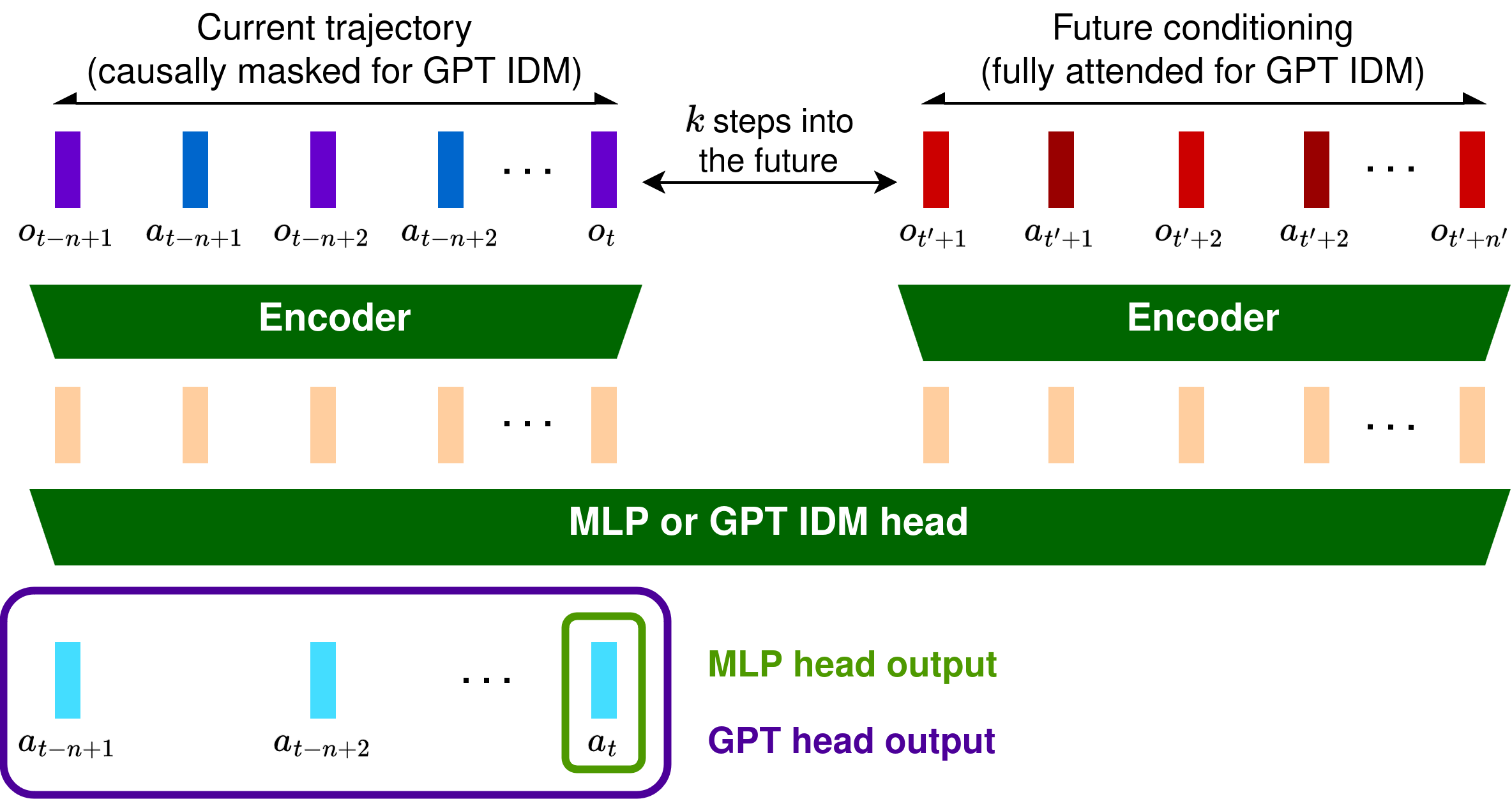}
    \caption{A high-level overview of the IDM model. We encode two distinct trajectories, the current trajectory of the agent, and the future conditioning. The resulting encodings are then passed into an IDM head to select which action should be performed.}
     \label{fig:models/idm_model}
\end{figure}

\begin{figure} [!t]
     \centering
     \includegraphics[width=0.78\textwidth]{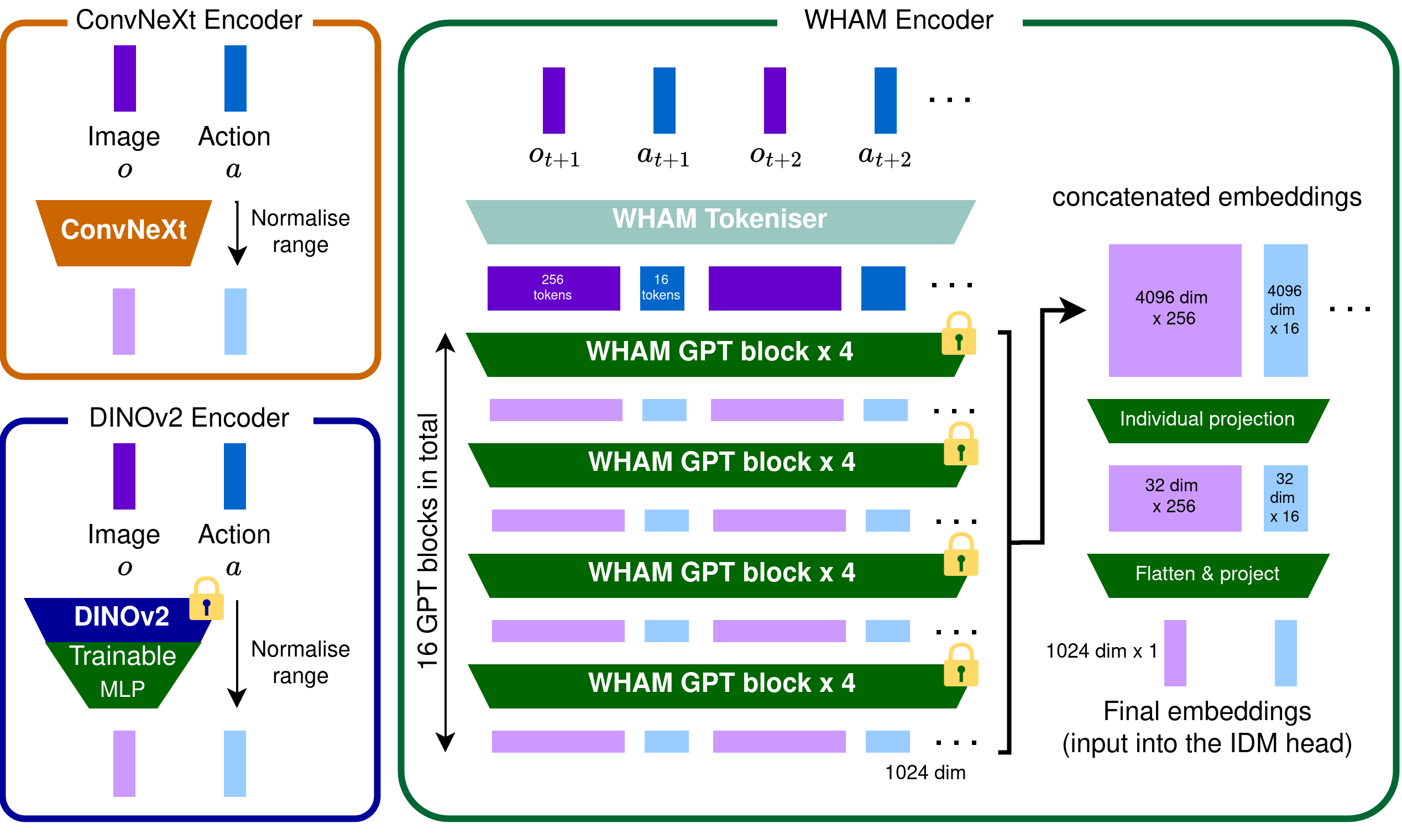}
    \caption{We evaluate three different encoders. A trained from scratch ConvNeXt encoder, a general pre-trained encoder DINOv2, and a game-specific pre-trained World and Human Action Model.}
     \label{fig:models/encoder_models}
\end{figure}

As shown in Figure~\ref{fig:models/idm_model}, our generalised IDM-$K$ model processes a past trajectory sequence up to time step $t$, and a future trajectory sequence from time step $t' + k$ onward. Given these inputs, the model predicts the action taken at $t$. Both observation and action sequences can be encoded independently or jointly, proving flexibility in representing past and future context. The encoded representations are then passed through a policy head to predict the action.

\paragraph{Visual Encoder}
We evaluate three different encoders (shown in \Cref{fig:models/encoder_models}):
\begin{enumerate}
    \item ConvNeXt~\citep{9879745}: A convolutional network trained from scratch, combined with a normalised action vector.
    \item DINOv2~\citep{oquab2024dinov}: A pre-trained general-purpose image encoder with a fine-tunable MLP head combined with a normalised action vector.
    \item World and Human Action Model (WHAM)~\citep{kanervisto2025world_wham}: A pre-trained game-specific encoder optimised for auto-regressive next-token prediction, processing sequences of interleaved image and action tokens as context. 
\end{enumerate}

The WHAM encoder presents a unique challenge due to its tokenised representation. In its original form, a single observation and corresponding action are represented by 256 and 16 tokens, respectively. This results in a sequence length 272 greater than the actual trajectory length. Preliminary experiments showed that training the IDM-$K$ to predict action vectors every 272 steps does not provide a sufficiently dense loss to the model. To address this, we project the sequential embeddings down to compact representations -- one for the observation input and another for the action input before passing them through a secondary projection layer to obtain the final embeddings.

\paragraph{Decoder} 
Since the model processes sequential data, we employ a transformer-based decoder for action prediction.
Specifically, We use a GPT model~\citep{radford_language_2019} as the IDM head to predict the actions taken to arrive at the future trajectory. With a GPT head, it is possible to predict multiple actions corresponding to different time steps at which the past information is cut off. These additional losses on the sequential action outputs help robustify the IDM policy to varying lengths of past trajectory input. We compare the GPT head against a Multi-Layer Perceptron (MLP) IDM head. The MLP head flattens and concatenates the input encodings, offering a simpler alternative to the transformer-based approach.

We combine the decoders (MLP, GPT), with the encoders (ConvNeXt, DINOv2, and WHAM) resulting in six IDM-$K$ variants to be evaluated.

\subsection{Future selection strategies} \label{sec:future_selection_strategies}

Selecting an appropriate future conditioning strategy for IDM-$K$ is crucial due to two primary challenges: (1) agent imperfections, and (2) environmental stochasticity.

\begin{itemize}
    \item Agent Imperfections: The model's action predictions are not always accurate, leading to deviations from the reference trajectory. As these errors accumulate, the agent may drift too far from the expected path. The distance between the current position and the future conditioning can become too large, putting the agent out of the training distribution.
    \item Environmental Stochasticity: Certain game elements, such as randomly spawning firewalls, moving platforms, and health packs can alter the environment unpredictably, causing a further distribution shift.
\end{itemize}

To mitigate these issues, we explore four future conditioning strategies:

\paragraph{Static future conditioning}
In this approach, the future start timestep $t'_{\text{future}}$ is determined solely by the current timestep $t_{\text{current}}$, independent of the agent's position. Given the number of frames to be skipped $K$, the future start timestep is computed as:
\begin{equation} 
    \label{eq:static_future}
    t'_{\text{future}} = t_{\text{current}} + K.
\end{equation}

This method does not account for spatial deviations, making it susceptible to distribution drift  when the agent strays from the intended trajectory.

\paragraph{Closest future conditioning}
This method selects the closest point in the reference trajectory based on the agent's current position, minimising spatial deviation. The future start timestep is given as:

\begin{equation}
    \label{eq:closest_future}
t'_{\text{future}} = \argmin_{t'}\lVert\tau(t') - \bm{\hat{x}}_t \rVert + K,
\end{equation}
where $\tau(t)$ represents the $(x, y, z)$ position in the reference trajectory at timestep $t$, $\bm{\hat{x}}_t$ represents the agent's current $(x, y, z)$ position, and $K$ represents the number of skipped frames.

While this strategy maintains spatial alignment, it introduces potential pitfalls in scenarios involving loops, or long stationary phases, where the agent may become trapped in an infinite cycle.

\paragraph{Radius future conditioning}
This method dynamically updates the future conditioning timestep based on the proximity to the reference trajectory. If the agent is within a predefined radius $r$, the future start timestep advances by one step, otherwise it remains unchanged. 
\begin{equation}
    \label{eq:radius_future}
    t'_{\text{future}} = \begin{cases} 
      t'_{\text{future}}  + 1, & \text{if}~ \lVert\tau(t'_{\text{future}}) - \bm{\hat{x}}_t\rVert \leq r, \\
      t'_{\text{future}}, & \text{otherwise}, \\
   \end{cases}
\end{equation}
where $\tau(t)$ represents the $(x, y, z)$ position in the reference trajectory at timestep $t$, $\bm{\hat{x}}_t$ represents the agent's current $(x, y, z)$ position, and $r$ represents the radius. 

\paragraph{Inner-Outer future conditioning}

This method introduces two thresholds: an inner radius $r_\text{in}$ and an outer radius $r_\text{out}$. The future timestep is updated based on the distance between the agent's position and the position of the current future conditioning frame.
\begin{itemize}
    \item Smaller than $r_\text{in}$ - The future timestep advances until it leaves the inner radius.
    \item Between $r_\text{in}$ and $r_\text{out}$ - The future advances by 1 step, same as the Radius strategy.
    \item Larger than $r_\text{out}$ - The future timestep remains unchanged.
\end{itemize}

A full description of the conditioning strategies can be seen in Appendix ~\ref{sec:future_selection_strategies_app}.
\section{Experiments} \label{sec:experiments}

The following section describes the environment used for the experiments, outlines the training and evaluation setup and provides and analyses the empirical results of the different models.

\subsection{Environment}

We evaluate our approach in Bleeding Edge\footnote{Official game website: \href{https://www.bleedingedge.com/en}{https://www.bleedingedge.com/en}}, a third-person multiplayer game featuring various playable characters with unique abilities. This environment provides a challenging testbed for trajectory-following agents due to its dynamic gameplay and stochastic elements.

\begin{figure} [!ht]
     \centering
     \begin{subfigure}[b]{0.45\textwidth}
         \centering
         \includegraphics[width=\textwidth]{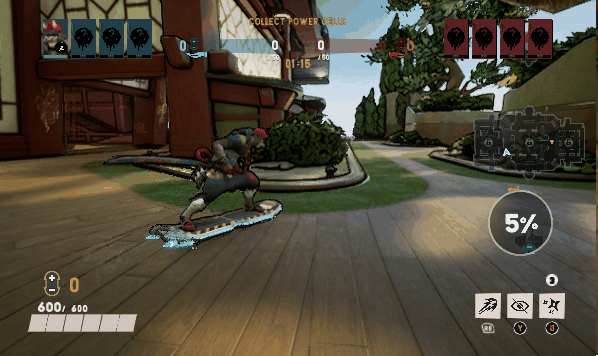}
         \caption{Sky Garden}
         \label{fig:sky_garden_map}
     \end{subfigure}
     \hfill
     \begin{subfigure}[b]{0.45\textwidth}
         \centering
         \includegraphics[width=\textwidth]{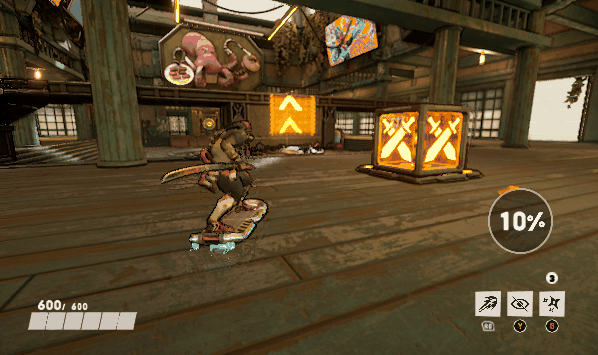}
         \caption{Dojo}
         \label{fig:dojo_map}
     \end{subfigure}
        \caption{Example images of the Sky Garden and Dojo maps used for training and evaluation.}
        \label{fig:maps}
\end{figure}

Our experiments focus on two maps: SkyGarden and the tutorial map Dojo (Figure~\ref{fig:maps}). The agent's observations contain both the visual signal, representing the agent's camera viewport and the symbolic modality including the telemetry data for the agent. The action space is represented by the input of an Xbox controller, having a discrete set of $12$ buttons and $2$ continuous sticks, for movement and camera rotation, each including the x and y axes.

The dataset contains $71,940$ trajectories of recorded human gameplay on the Sky Garden map, from 8788 matches recorded between 02-09-2020 and 19-10-2022~\citep{jelley2024aligningagentslikelarge}. This dataset represents approximately 1.12 years of human gameplay, and contains visual and telemetry data, as well as the actions the players took during the game.  
\subsection{Setup} 

To unify the modality of the action space from the dataset, we normalise the continuous dimensions (stick inputs) of the continuous action space into a [-1, 1] range and then discretise it into $11$ bins. All of the proposed models were trained for 200 epochs, each epoch with 1,000 updates with a batch size of 64.
The complete list of training hyperparameters is provided in Appendix~\ref{sec:training_hyperparams}.

\paragraph{Evaluation} 
We evaluate model performance on 8 held-out trajectories, covering diverse tasks:
\begin{itemize}
    \item Jumppad (3 trajectories): The agent must select the correct path at a crossroads
    \item Benchmark (3 trajectories): Involves complex turns throughout different map areas.
    \item Dojo (2 trajectories): Navigation through a tutorial environment.
\end{itemize}

These test trajectories vary in complexity and domain. Some originate from Sky Garden (the primary testing environment), while others come from Dojo, introducing an additional domain shift. The trajectories can be seen at \href{https://adaptingworldmodels.github.io/}{https://adaptingworldmodels.github.io/}.

Each agent is evaluated on 10 rollout seeds per trajectory. We assess the performance on two metrics:
\begin{itemize}
    \item AUC (Area Under the Curve): Measures trajectory similarity via Dynamic Time Warping (DTW)~\citep{Müller2007} at varying radii.
    \item Future Index Ratio (FI): Captures the proportion of the trajectory followed before the first significant deviation from the reference path.
\end{itemize}

 A detailed formulation for the AUC metric is provided in Appendix~\ref{sec:auc}. Unless otherwise stated, all reported results reflect the median score across 10 rollouts per model, per trajectory.

\subsection{Results}

This section contains the empirical evaluation of the models and provides insight into the encoder and IDM head design choices. We split the section into two parts: (1) Evaluation in the General setting, and (2) Evaluation of Specific and Fine-tuning settings.

\subsubsection{General Setting}

Table~\ref{tab:zero-shot} summarises the performance of different architectures trained on a large, diverse dataset and evaluated zero-shot on unseen trajectories. Across all evaluated architectures, ConvNeXt outperforms the other encoders, regardless of whether paired with the MLP or GPT IDM head. The FI performance of the models is available in Appendix~\ref{tab:full_f1_results_general}. While the AUC provides a quantitative comparison between models, it does not capture if trajectory-following was successful.\footnote{Sample videos of the ConvNeXt agent are available at \href{https://adaptingworldmodels.github.io/}{https://adaptingworldmodels.github.io/}} 

\begin{table}[!ht]
    \caption{AUC for the general setting evaluation on 8 heldout trajectories. Higher AUC is better.}
    \label{tab:zero-shot}
    \centering
    \begin{tabular}{lcccccc}
        \toprule
        & \multicolumn{3}{c}{MLP} & \multicolumn{3}{c}{GPT} \\
        \cmidrule(lr){2-4}\cmidrule(lr){5-7}
        Trajectory & ConvNeXt & DINOv2 & WHAM & ConvNeXt & DINOv2 & WHAM \\
        \midrule
        Jumppad left    & 0.99 & 0.92 & 0.84 & \textbf{0.99} & \textbf{0.99}          & 0.87 \\
        Jumppad right   & 0.89 & 0.88 & 0.91 & \textbf{0.99} & \textbf{0.99}          & 0.92 \\
        Jumppad mid     & 0.99 & 0.81 & 0.83 & \textbf{0.99} & \textbf{0.99}          & 0.90 \\
        Benchmark 0     & 0.33 & 0.33 & 0.17 & \textbf{0.64} & 0.59          & 0.29 \\
        Benchmark 1     & 0.81 & 0.50 & 0.67 & \textbf{0.98} & 0.88          & 0.57 \\
        Benchmark 2     & 0.73 & 0.56 & 0.26 & \textbf{0.95} & 0.89          &	0.50 \\
        Dojo Ramp      & 0.65 & 0.40 & 0.67 & \textbf{0.69}          & \textbf{0.69} &	0.68 \\
        Dojo Gong       & 0.54 & 0.42 & 0.39 & \textbf{0.66} & 0.65          &	0.64 \\
        Mean            & 0.74 & 0.60 & 0.59 & \textbf{0.86} & 0.84          & 0.67 \\
        \bottomrule
    \end{tabular}
\end{table}

\begin{figure}
    \begin{subfigure}{\textwidth}
    {
    \begin{minipage}{\linewidth}
    \includegraphics[width=.24\linewidth]{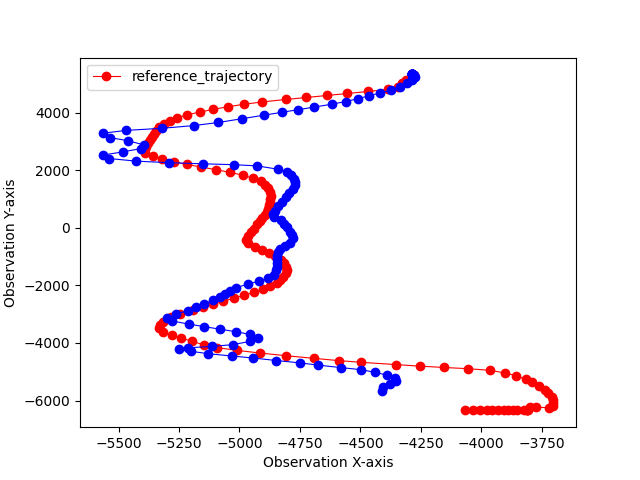}\hfill
    \includegraphics[width=.24\linewidth]{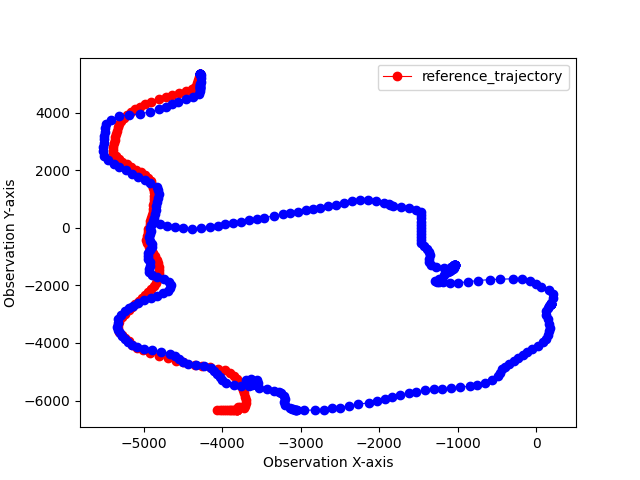}\hfill
    \includegraphics[width=.24\linewidth]{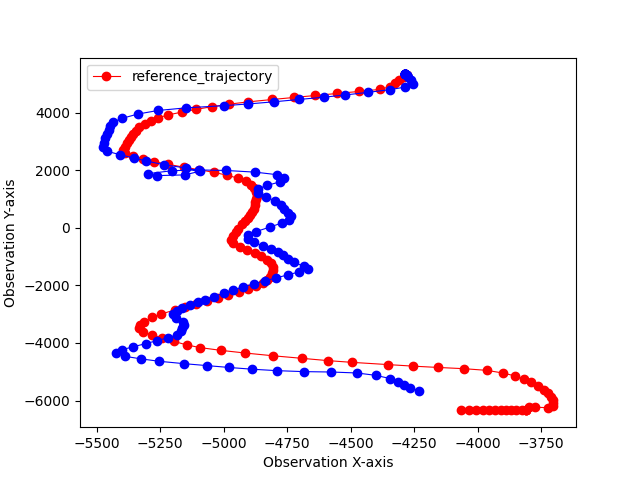}\hfill
    \includegraphics[width=.24\linewidth]{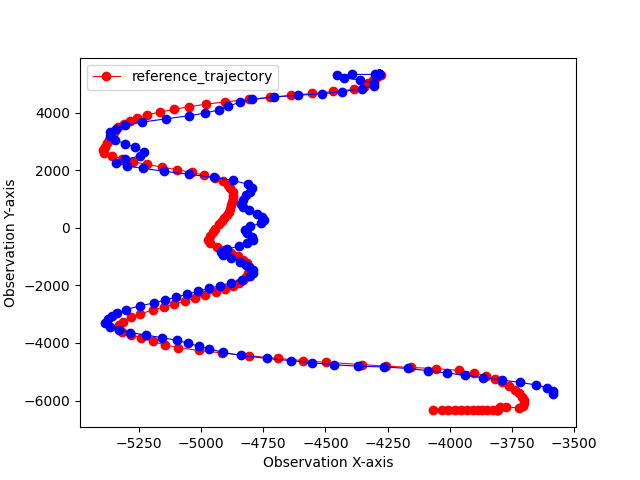}
    \caption{Samples of ConvNeXt-GPT rollouts on the Benchmark 1 trajectory.}
    \label{fig:cgpt-trajectories}
    \end{minipage}
    }
    \end{subfigure}
    \begin{subfigure}{\textwidth}
    {
    \begin{minipage}{\linewidth}
    \includegraphics[width=.24\linewidth]{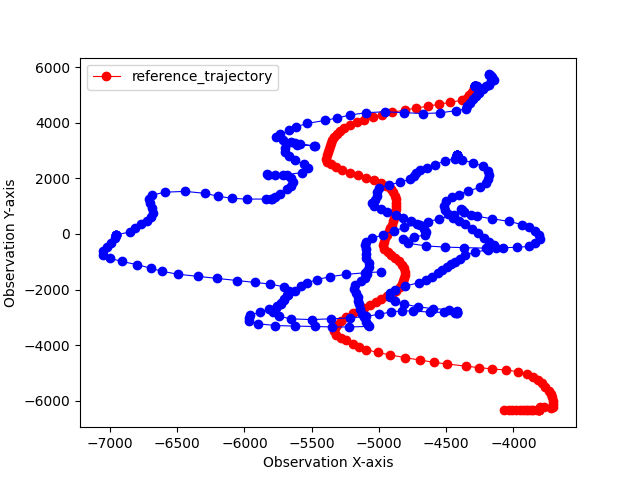}\hfill
    \includegraphics[width=.24\linewidth]{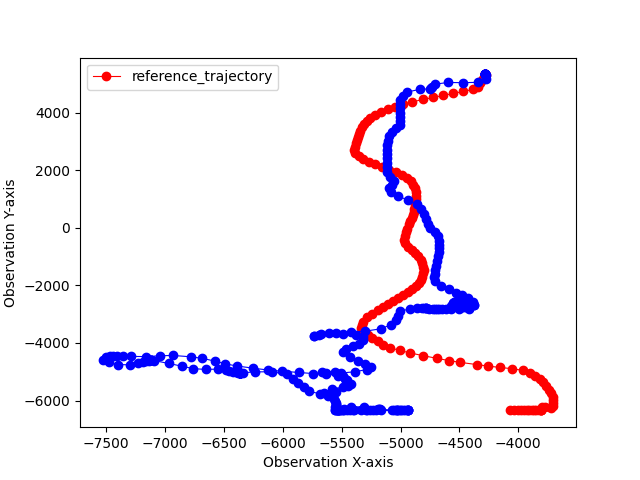}\hfill
    \includegraphics[width=.24\linewidth]{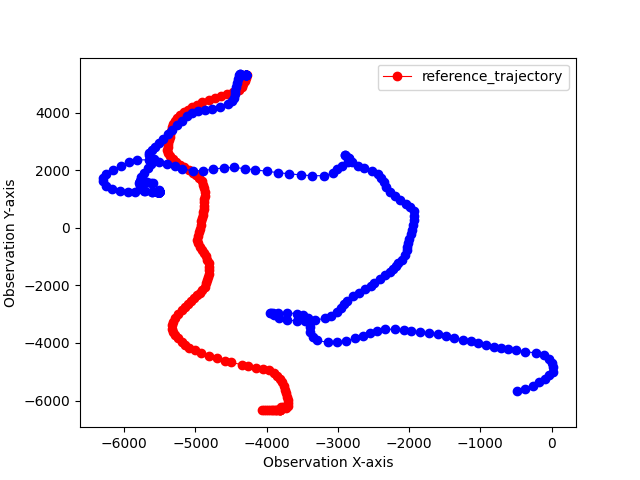}\hfill
    \includegraphics[width=.24\linewidth]{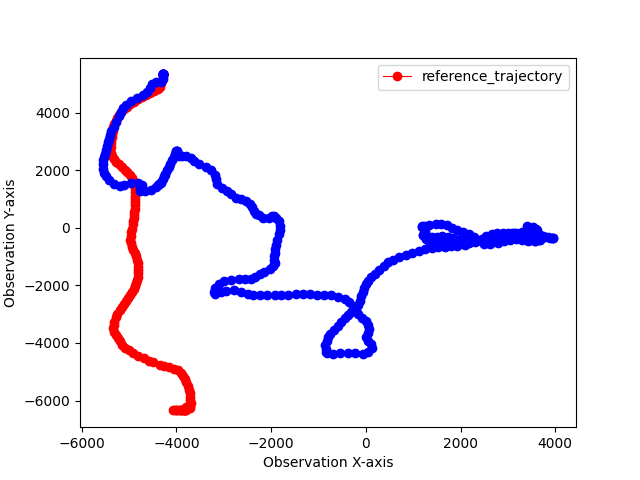}
    \caption{Samples of DINOv2-GPT rollouts on the Benchmark 1 trajectory.}
     \label{fig:dgpt-trajectories}
    \end{minipage}
    }
    \end{subfigure}
    \begin{subfigure}{\textwidth}
    {
    \begin{minipage}{\linewidth}
    \includegraphics[width=.24\linewidth]{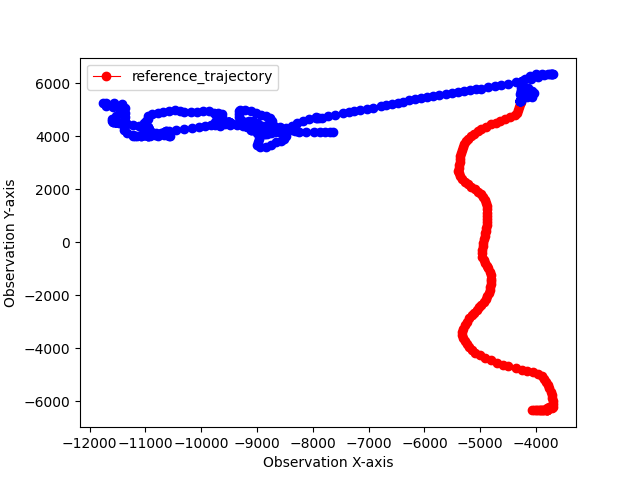}\hfill
    \includegraphics[width=.24\linewidth]{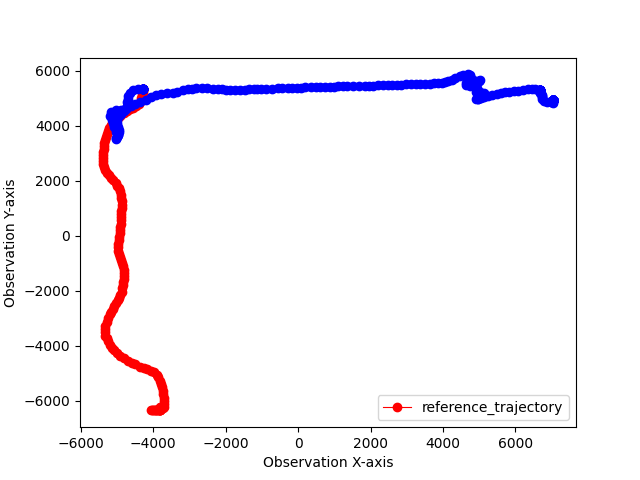}\hfill
    \includegraphics[width=.24\linewidth]{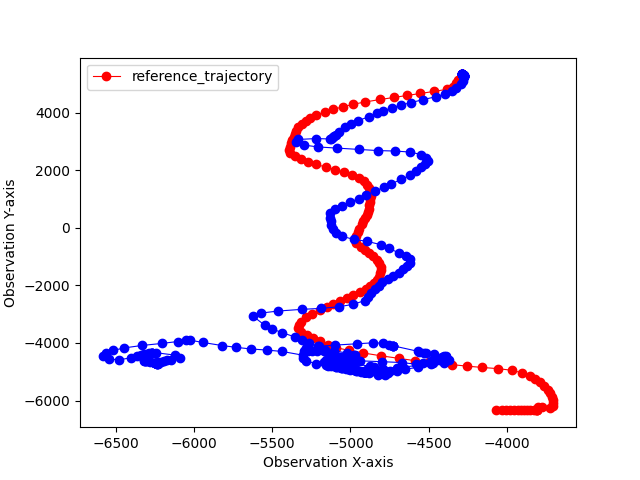}\hfill
    \includegraphics[width=.24\linewidth]{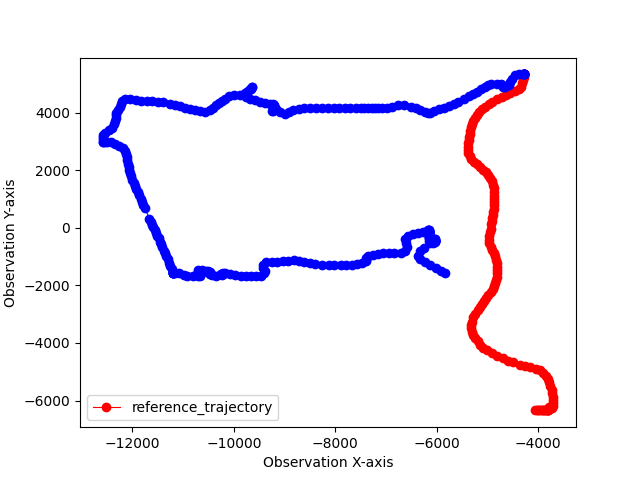}
    \caption{Samples of WHAM-GPT rollouts on the Benchmark 1 trajectory.}
     \label{fig:wgpt-trajectories}
    \end{minipage}
    }
    \end{subfigure}
\caption{Sampled rollouts from Benchmark 1. The red line shows the reference trajectory, while the blue line shows the agent's path. $x$ and $y$ axes represent the coordinates of the agent.}
\label{fig:rollout-trajectories}
\end{figure}

To provide a more intuitive understanding of model performance, Figure~\ref{fig:rollout-trajectories} visualises the Benchmark 1 trajectory, overlaying the agent's paths to the reference trajectory.
Figure~\ref{fig:rollout-trajectories} reveals distinct performance differences among the models. ConvNeXt consistently follows the expected trajectory, with deviations only in a single rollout. DINOv2 exhibits occasional successful runs, recovering from minor deviations, while WHAM fails to follow the trajectory, displaying mostly arbitrary movement. These qualitative observations align will the quantitative AUC results presented in Table 1.

Our results suggest that training an encoder specifically for trajectory following is advantageous. Even the relatively simple ConvNeXt architecture produces more effective embeddings than large-scale pre-trained models like DINOv2 and WHAM, which were not explicitly trained for this task.

The performance difference between ConvNeXt, DINOv2 and WHAM likely comes from differences in their embedding strategies. ConvNeXt and DINOv2 encode observations independently, with actions processed separately, while WHAM uses a causal transformer to jointly encode both modalities. This early-stage entanglement of observations and actions may limit the model's ability to generalise effectively to trajectory-following tasks.

While WHAM-GPT achieves lower training loss compared to ConvNeXt and DINOv2, this advantage does not translate to better evaluation performance, where WHAM underperforms relative to other variants. This discrepancy likely originates from the difference between training and evaluation conditions. During training, past and future trajectories are perfectly aligned both spatially and temporally, as they originate from the same recorded trajectory. However, at evaluation time, past trajectories are generated dynamically by the agent, while future trajectories remain fixed from recording. The training curves for different models can be seen in Appendix~\ref{app:training_curves}.

\paragraph{General Setting Ablations}

To understand the benefit of having different input modalities, we conduct ablations by selectively removing observations or actions. Table~\ref{tab:different-input-ablations} presents the performance comparison for the ConvNeXt-GPT model across these variations. 

Results indicate that visual inputs are essential for accurate trajectory following, while including action inputs provides only marginal gains. In simple scenarios (e.g. Jumppad trajectories), action inputs alone suffice. The observation-only model almost matches the performance of the full model.
The evaluation shows that having the visual modality is crucial while adding the action modality gives only a slight overall improvement. Full results per trajectory can be seen in Appendix~\ref{sec:app:full-results}.

\begin{table}[!ht]
    \caption{AUC and the FI of different inputs to the model - Mean results across all trajectories}
    \label{tab:different-input-ablations}
    \centering
    \begin{tabular}{lcccccc}
        \toprule
          &  \multicolumn{2}{c}{Observations Only} & \multicolumn{2}{c}{Actions Only} & \multicolumn{2}{c}{Full Model} \\
        \cmidrule(lr){2-3}\cmidrule(lr){4-5}\cmidrule(lr){6-7}
              & AUC  & FI   & AUC  & FI   & AUC  & FI   \\
        \midrule
        Mean & 0.84 & 0.70 & 0.69 & 0.47 & \textbf{0.86} & \textbf{0.73} \\
        \bottomrule
    \end{tabular}
\end{table}

We also investigate the impact of sequence length on model performance. Table~\ref{tab:future-past-length-performance} reports results for different past and future sequence lengths, alongside a BC agent as a baseline (which lacks future conditioning by design).
Full results per trajectory can be seen in Appendix~\ref{sec:app:full-results}.

The results indicate that the BC agent struggles to complete any of the trajectories consistently; the only trajectory it performed well on was Jumppad mid, where the behaviour is to only move forward. Other models performed on par with each other, with giving full 10 observation-action frames for both the past and the future being slightly better. Full results per trajectory are in Appendix~\ref{sec:app:full-results}.

\begin{table}[!ht]
    \caption{AUC and the FI of different future and past lengths - Mean results across all trajectories}
    \label{tab:future-past-length-performance}
    \centering
        \begin{tabular}{lcccccccccc}
        \toprule
              & \multicolumn{2}{c}{BC}   & \multicolumn{2}{c}{1P-1F}   & \multicolumn{2}{c}{10P-1F}   & \multicolumn{2}{c}{1P-10F} & \multicolumn{2}{c}{Full model}    \\
        \cmidrule(lr){2-3}\cmidrule(lr){4-5}\cmidrule(lr){6-7}\cmidrule(lr){8-9}\cmidrule(lr){10-11}
         & AUC  & FI   & AUC  & FI   & AUC  & FI   & AUC  & FI   & AUC  & FI   \\
        \midrule
        Mean & 0.64 & 0.45 & 0.84 & 0.69 & 0.85 & 0.69 & 0.79 & 0.64 & \textbf{ 0.86} &\textbf{ 0.73} \\
        \bottomrule
    \end{tabular}
\end{table}

We present the results for the future selection strategies in Table~\ref{tab:future-selection-strategies}. It is important to note we do not measure FI in this setting, as the radius values directly influence the allowed spatial distance between the agent and the future trajectory, making a fair assessment in this regard impractical. Among the four selection strategies \textit{Closest} slightly outperforms the others, achieving the AUC of $0.877$. While \textit{Closest} was the best-performing strategy on the benchmark, we chose the \textit{Radius} strategy for our models for two key reasons. First, the agent struggled to follow the Dojo trajectories, showing arbitrary movement regardless of the strategy. Excluding the Dojo trajectories, the \textit{Radius} strategy yields a slightly better mean AUC than \textit{Closest}. Second, the \textit{Closest} strategy encounters theoretical issues when dealing with trajectories containing loops or extended sequences of no-ops - problems that were not present in the evaluation set, but may arise in other scenarios.

\begin{table}[!ht]
    \caption{Different future selection strategies. Results show the median AUC value across all trajectories. The radius $R$ for Radius and Inner-Outer strategies has been determined through a parameter sweep for each trajectory.}
    \label{tab:future-selection-strategies}
    \centering
    \begin{tabular}{lcccc}
        \toprule
        & Static    & Closest   & Radius & Inner-Outer \\
        \midrule
        Mean AUC           & 0.84 &   \textbf{0.88}	&    0.86   &	0.85 \\
        \bottomrule
    \end{tabular}
\end{table}

\subsubsection{Specific and Fine-tuning}

We trained models on 30 trajectories that exhibit near-identical behaviour (variations of Dojo Ramp). Table~\ref{tab:specific-and-finetuned-results} presents the results. The evaluation procedure matches the General setting, the only difference being that the WHAM encoder was not evaluated. The reason for this is that the WHAM model requires the availability of the large dataset that these experiments assume is not available.

\begin{table}[!ht]
    \caption{AUC and FI in Dojo Ramp for models trained on the whole dataset and evaluated zero-shot on Dojo Ramp \textbf{(general training)}, models trained only on Dojo Ramp \textbf{(specific training)}, and models first trained on the whole dataset and fine-tuned on Dojo Ramp \textbf{(fine-tuning)}.}
    \label{tab:specific-and-finetuned-results}
    \centering
        \begin{tabular}{lcccccccc}
        \toprule
              & \multicolumn{2}{c}{ConvNeXt-MLP}   & \multicolumn{2}{c}{DINO-MLP}   & \multicolumn{2}{c}{ConvNeXt-GPT}   & \multicolumn{2}{c}{DINO-GPT}    \\
        \cmidrule(lr){2-3}\cmidrule(lr){4-5}\cmidrule(lr){6-7}\cmidrule(lr){8-9}
        & AUC  & FI   & AUC  & FI   & AUC  & FI   & AUC  & FI \\
        \midrule
        General training    & 0.65 & 0.22 & 0.40 & 0.21 & \textbf{0.69} & \textbf{0.29} & \textbf{0.69} & 0.24 \\
        Specific training   & 0.70 & 0.26 & 0.96 & \textbf{1.00} & 0.94 & 0.98 & \textbf{0.97} & \textbf{1.00} \\
        Fine-tuning          & \textbf{0.96} & \textbf{1.00} & 0.83 & 0.67 & \textbf{0.96} & \textbf{1.00} & 0.93 & 0.92 \\
        \bottomrule
    \end{tabular}
\end{table}

In the Specific setting, DINOv2-GPT achieves the best overall performance, closely followed by DINOv2-MLP. Both models achieve a perfect FI of $1.00$, indicating a replicated trajectory. 

Additionally, we would like to note the overall difference that the training on the specific behaviour makes for the evaluation on that behaviour. In the general setting, when we evaluated the agents on the previously unseen Dojo Ramp behaviour, none of the models could follow the trajectory, while when trained on the specific, Dojo Ramp trajectory, almost all were able to capture the full behaviour.

For the fine-tuning setting, we see that the ConvNeXt-MLP, ConvNeXt-GPT and DINOv2-GPT are all able to successfully follow the trajectory, showing that the ConvNeXt architecture, due to being trained from scratch benefits from the large initial dataset from the general setting, and ultimately slightly outperforming the DINOv2 models.

We also present the impact of the input sequence length to understand the impact of future conditioning in this setting. The results in Table~\ref{tab:specific-future-past-length-performance} show that even the BC version of the DINOv2-GPT model, where we don't condition the agent at all, still performs optimally. 
This is due to the nature of the training and evaluation setting. For the specific setting, there isn't any variety in the exhibited behaviour, where all trajectories follow the same path. In such a scenario, knowing the past is sufficient to predict the next action.

\begin{table}[!ht]
    \caption{AUC and FI for different past and future lengths, for Specific trajectories.}
    \label{tab:specific-future-past-length-performance}
    \centering
        \begin{tabular}{lcccccccc}
        \toprule
              & \multicolumn{2}{c}{ConvNeXt-MLP}   & \multicolumn{2}{c}{DINO-MLP}   & \multicolumn{2}{c}{ConvNeXt-GPT}   & \multicolumn{2}{c}{DINO-GPT}    \\
        \cmidrule(lr){2-3}\cmidrule(lr){4-5}\cmidrule(lr){6-7}\cmidrule(lr){8-9}
        & AUC & FI  & AUC & FI  & AUC & FI  & AUC & FI \\
        \midrule
        10P-0F (BC) & 0.70 & 0.26 & 0.96 & \textbf{1.00} &\textbf{ 0.97} & \textbf{1.00} & 0.93 & 0.99 \\
        10P-1F      & 0.71 & 0.26 & 0.74 & 0.85 & 0.96 & \textbf{1.00} & \textbf{0.98} & \textbf{1.00} \\
        10P-10F     & 0.70 & 0.26 & 0.96 & \textbf{1.00} & 0.94 & 0.98 & \textbf{0.97} & \textbf{1.00} \\
        \bottomrule
    \end{tabular}
\end{table}

\section{Conclusion} \label{sec:conclusion}

We evaluated different IDM architectures across three settings -- General, Specific, and Finetuned -- finding that different architectures excel in different scenarios. ConvNeXt-GPT performed best in the General setting, DINOv2 variants in the Specific setting, and ConvNeXt variants in the Finetuned setting. Additionally, we conducted extensive ablations to identify key design choices for inverse dynamics models in imitation learning. Additionally, we explored various future selection strategies to mitigate distribution shifts, a crucial challenge in trajectory-following tasks.

Despite the advancements in model architectures and design choices, trajectory-following remains a significant challenge. In the General setting, none of the agents successfully followed the most complex trajectories, highlighting persistent gaps in generalisation across different behaviours.
\paragraph{Limitations and Future Work}

We plan on further investigating model robustness against external perturbations, such as human intervention, by manually rotating the camera and inputting specific actions, to better understand recovery from out-of-distribution situations.

While our models were trained on all 13 Bleeding Edge characters and actions, our evaluation focused on movement trajectories, with a single character. Expanding testing to include behaviours such as combat interactions and specific character actions could provide a deeper insight into model capabilities and limitations.

\section*{Acknowledgements}
The authors would like to thank Katja Hofmann, Tabish Rashid, Tim Pearce, Yuhan Cao, Chentian Jiang, Shanzheng Tan, and Linda Yilin Wen at the Microsoft Research Game Intelligence Team for their immense support and contribution throughout this research.

\bibliography{main.bib}
\bibliographystyle{iclr2025_conference}

\newpage 

\appendix

\renewcommand\thefigure{\thesection.\arabic{figure}}  
\setcounter{figure}{0}
\setcounter{table}{0}
\renewcommand{\thetable}{\thesection.\arabic{table}}
\section{Appendix} \label{sec:appendix}

\subsection{Data modality}

The dataset consists of 8,788 recorded matches spanning from September 2, 2020, to October 19, 2022. In total, these trajectories represent approximately 1.12 years of cumulative human gameplay. The original 60Hz data has been downsampled to 10Hz. Each trajectory includes both visual and telemetry data, along with player actions taken during gameplay.

The resolution of the visual modality in the dataset is $600\times360$ which is then down-sampled to $128\times128$ before passing it to the encoder. 

\begin{figure}[!ht]
    \centering
    \includegraphics[width=0.5\linewidth]{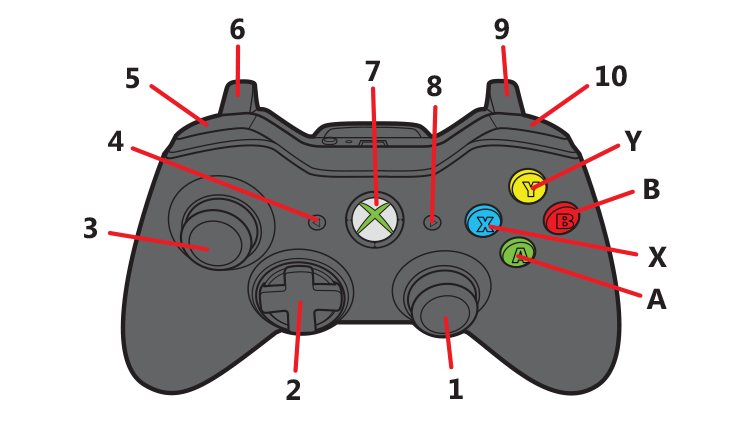}
    \caption{
        Xbox controller input. Labels $1$ and $3$ represent the continuous stick inputs - each stick has two axes it can move in, while other labels represent the discrete button inputs.
        }
    \label{fig:xbox_controller_input}
\end{figure}

For the action space, we used the standard Xbox controller scheme. Xbox controller has continuous input through the movement sticks, and discrete inputs through buttons, as seen in Figure~\ref{fig:xbox_controller_input}.\footnote{Controller image taken from: \href{https://support.xbox.com/en-US/help/xbox-360/accessories/controllers}{https://support.xbox.com/en-US/help/xbox-360/accessories/controllers}} 
We normalise the continuous action space into a [-1, 1] range, discretise it into 11 bins, and treat all buttons as discrete values. 

\subsection{Training} \label{sec:training}

\subsubsection{hyperparameters} \label{sec:training_hyperparams}

To make the various MLP and GPT models comparable, all latent dimensions were set to 1024 for the ConvNeXt~\citep{9879745}, DINOv2~\citep{oquab2024dinov} and WHAM~\citep{kanervisto2025world_wham} encoders. 4 layers with hidden dimensions of 1024 each were used for the MLP IDMs, whereas 4 GPT blocks with hidden dimensions of 1024 each were used for the GPT IDMs. In addition, 4 MLP layers with hidden dimensions of 1024 each were used as learnable state encoder layers for the pre-trained DINOv2 encoder. We trained on a single NVIDIA H100 80GB GPU machine. The learning rate used was $0.0001$.

\subsubsection{Training Curves}
\label{app:training_curves}

\begin{figure} [!h]
     \centering
     \includegraphics[width=0.83\textwidth]{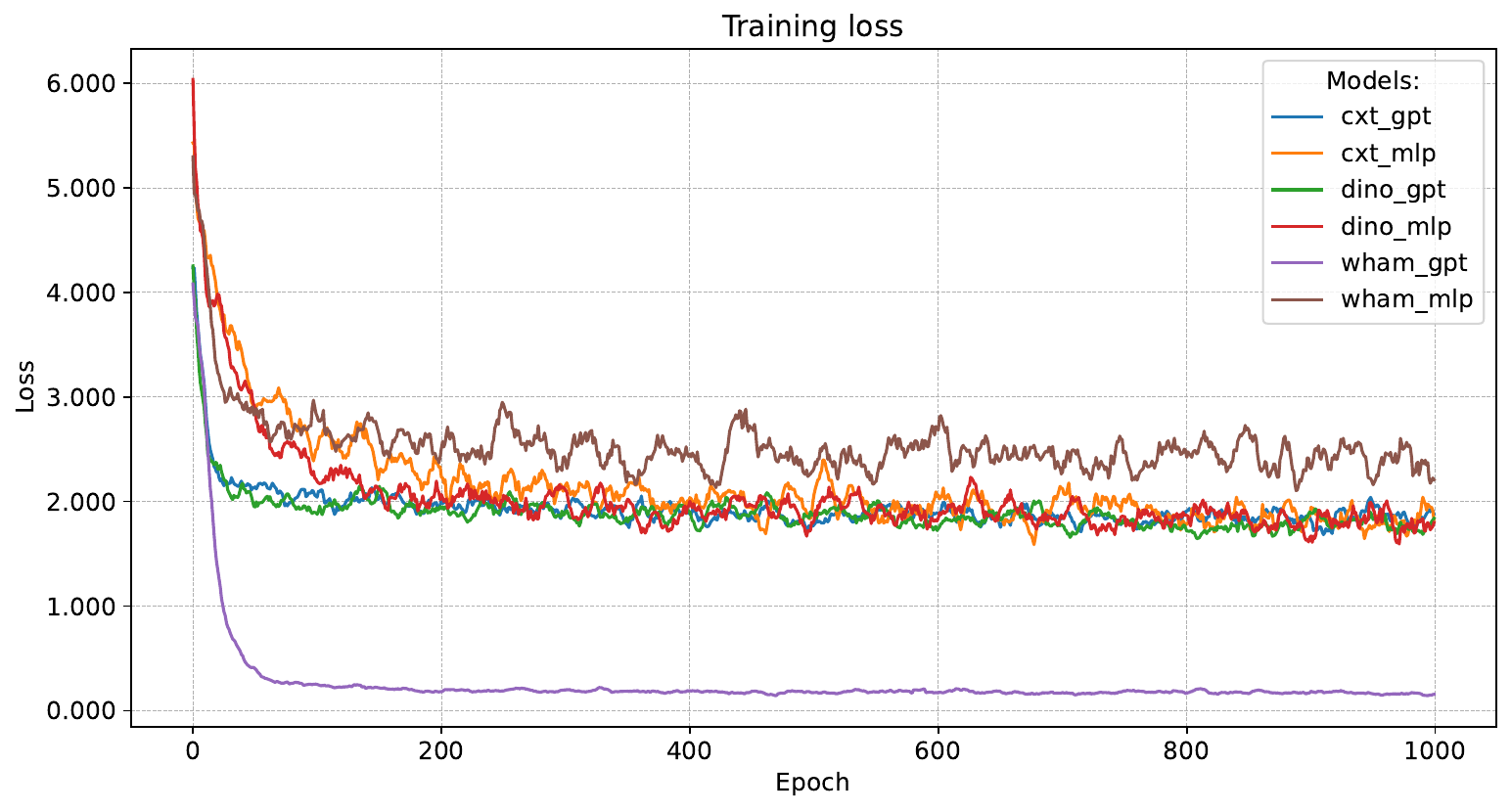}
    \caption{Training loss curve.}
     \label{fig:training_loss_general}
\end{figure}

\begin{figure} [!h]
     \centering
     \includegraphics[width=0.83\textwidth]{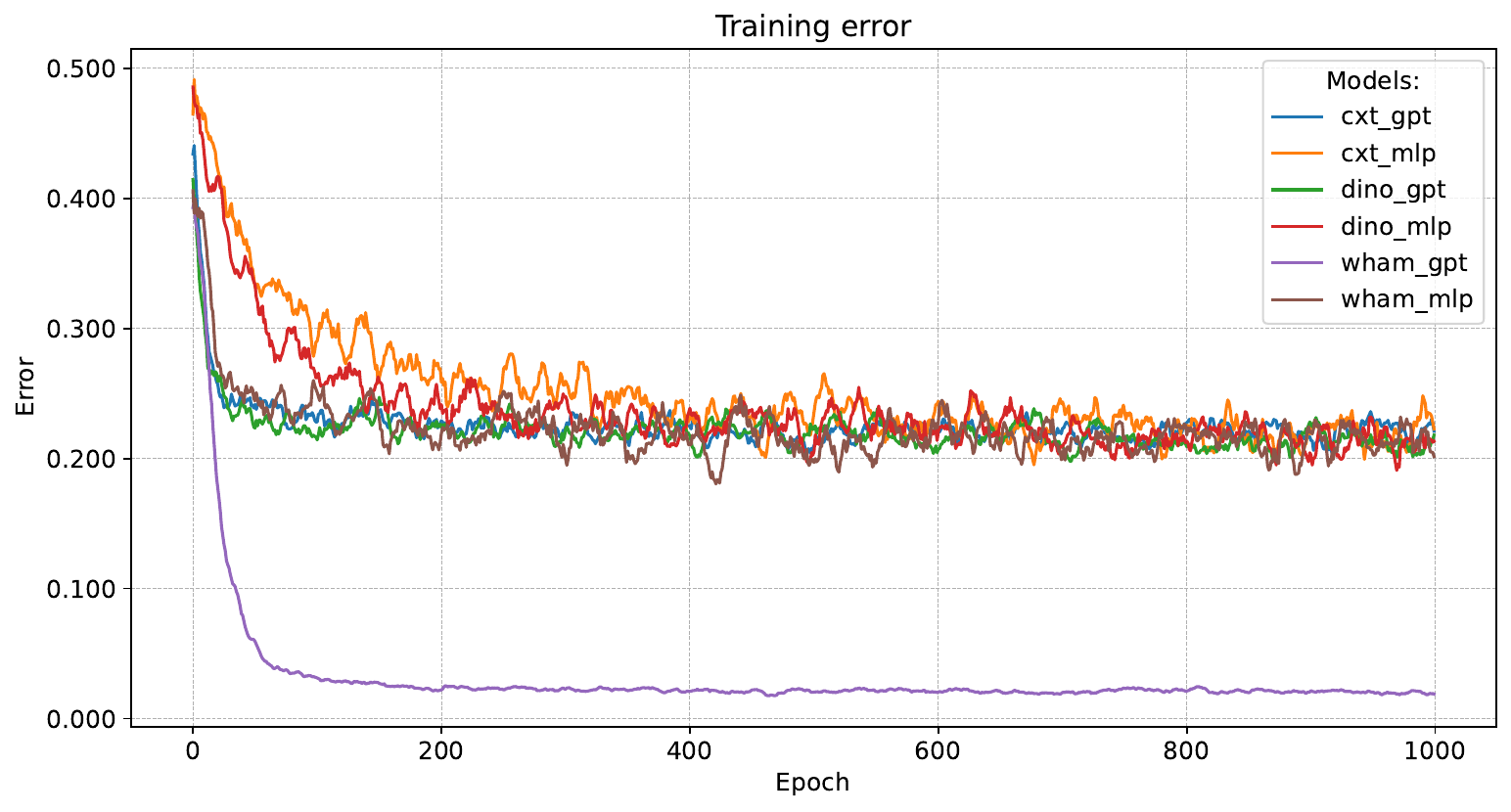}
    \caption{Training error curve.}
     \label{fig:training_error_general}
\end{figure}

\begin{figure} [!h]
     \centering
     \includegraphics[width=0.83\textwidth]{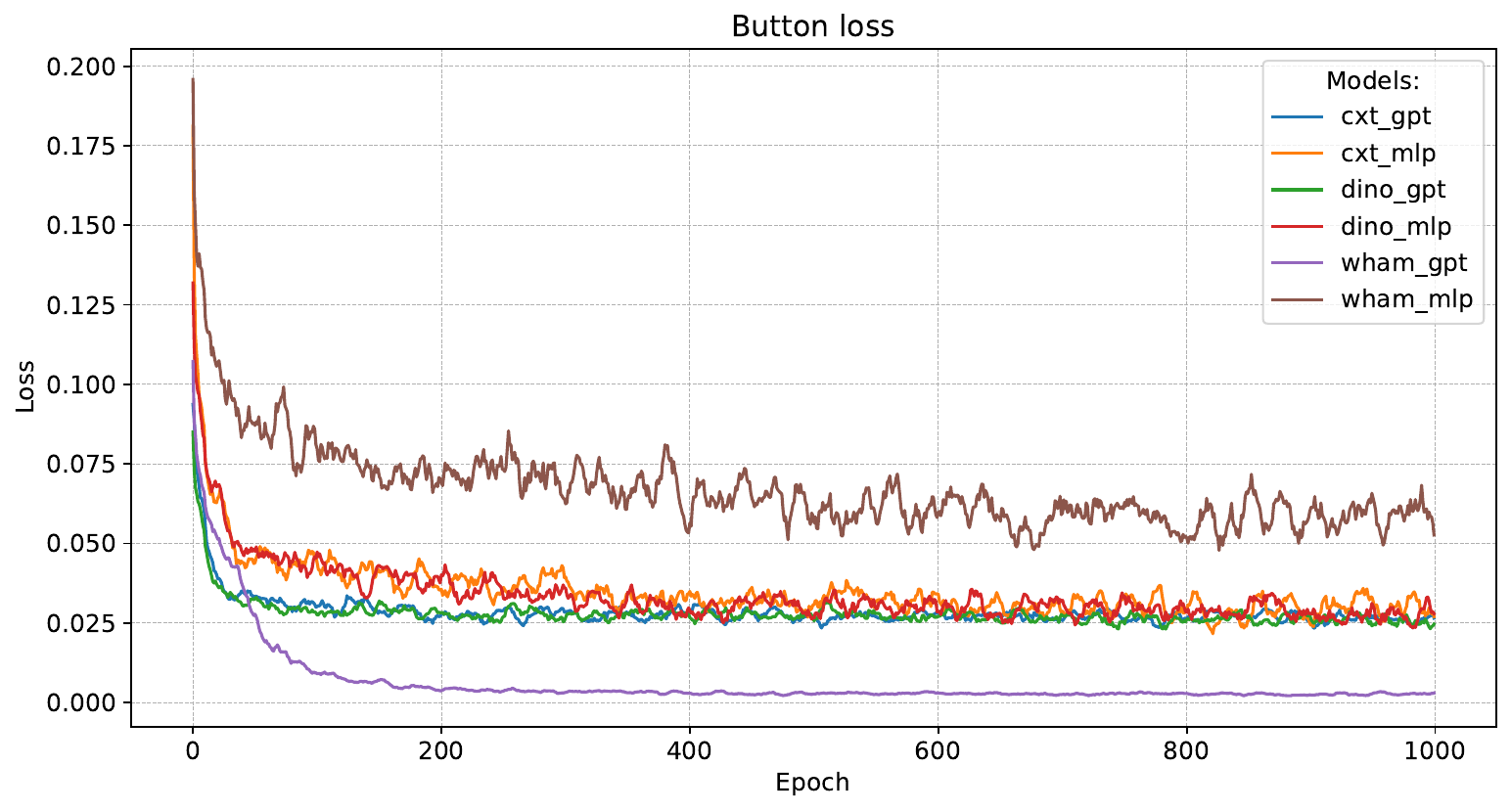}
    \caption{Button loss curve for the six model variants, trained in the general setting.}
     \label{fig:button_loss_general}
\end{figure}

\begin{figure} [!h]
     \centering
     \includegraphics[width=0.83\textwidth]{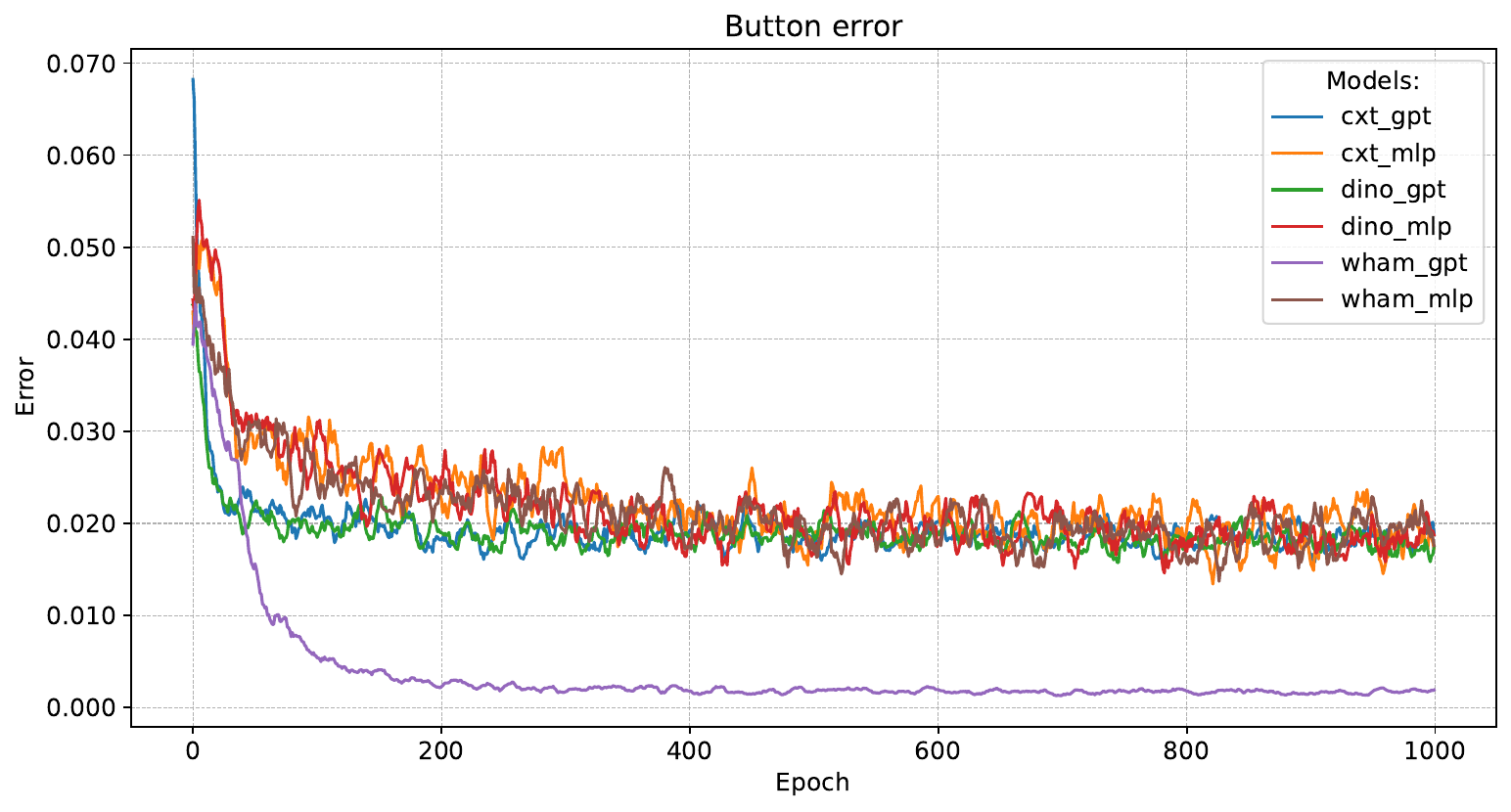}
    \caption{Button error curve for the six model variants, trained in the general setting.}
     \label{fig:button_error_general}
\end{figure}

\begin{figure} [!h]
     \centering
     \includegraphics[width=0.83\textwidth]{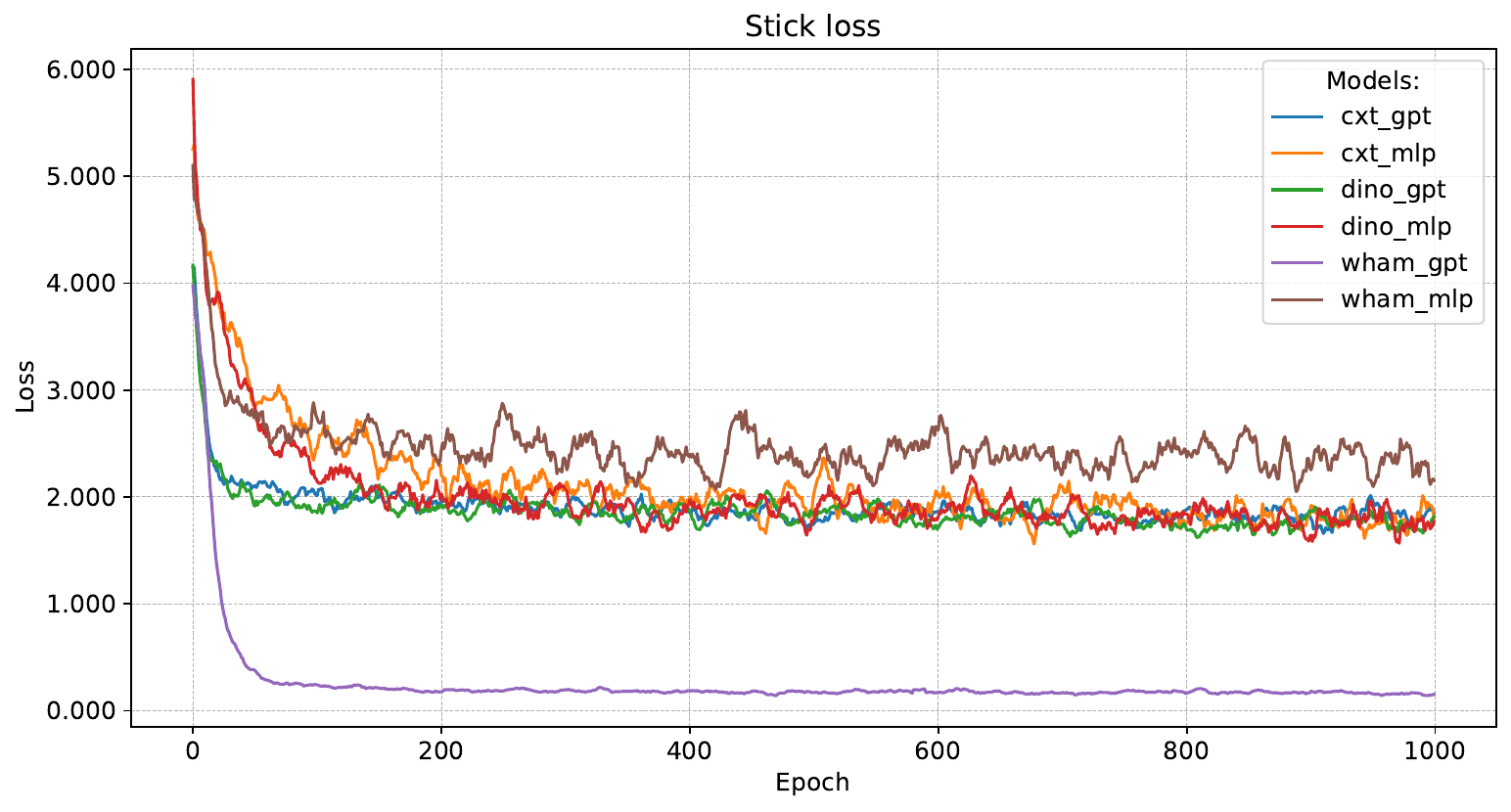}
    \caption{Sticks loss curve for the six model variants, trained in the general setting.}
     \label{fig:sticks_loss_general}
\end{figure}

\begin{figure} [!h]
     \centering
     \includegraphics[width=0.83\textwidth]{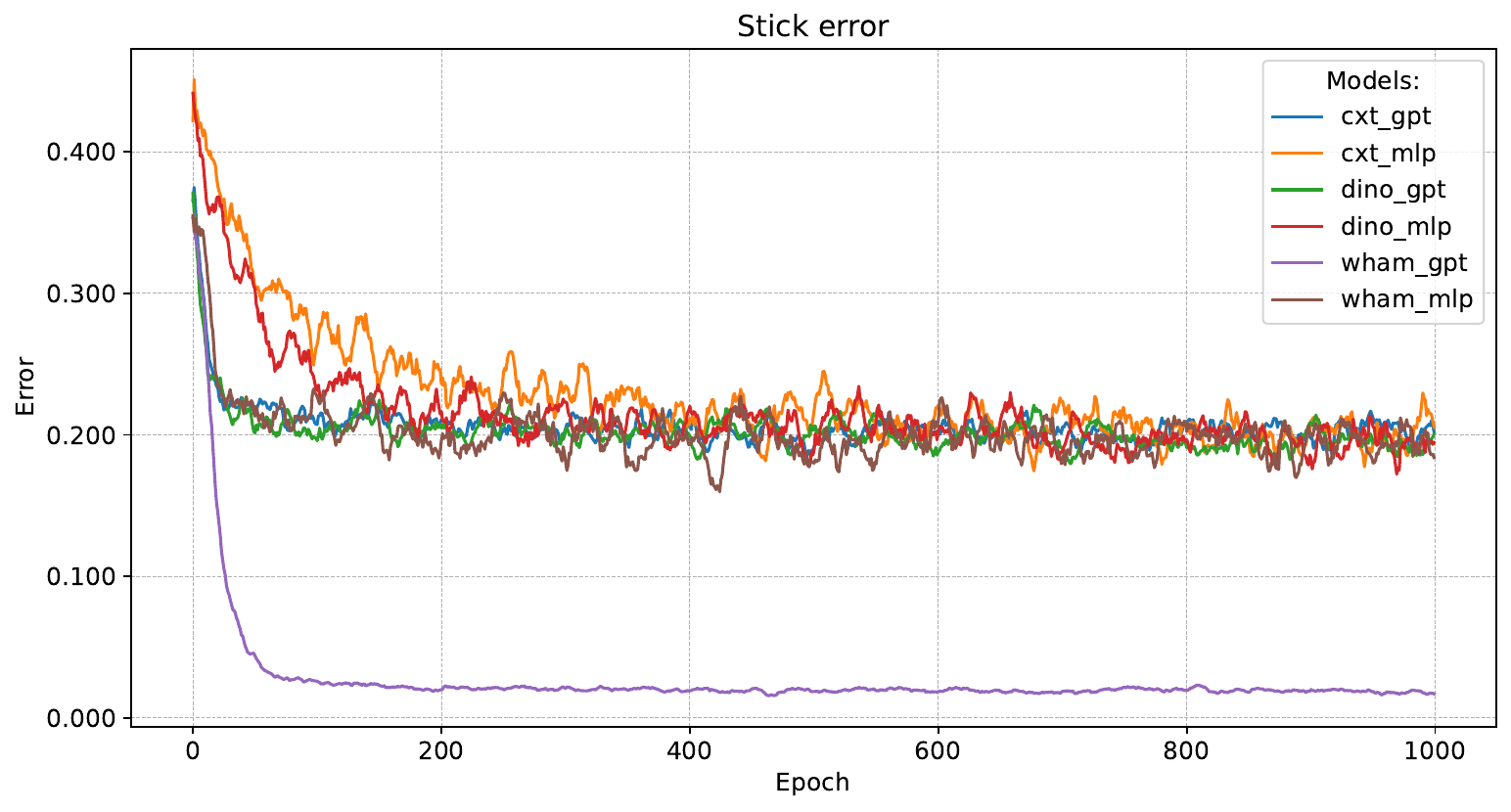}
    \caption{Sticks error curve for the six model variants, trained in the general setting.}
     \label{fig:sticks_error_general}
\end{figure}

\Cref{fig:training_loss_general,fig:training_error_general,fig:button_loss_general,fig:button_error_general,fig:sticks_loss_general,fig:sticks_error_general} present the training curves of the model with separated continuous and discrete modalities, to further investigate the capability of the model for each action space.

\newpage

\subsection{Future selection strategies}
\label{sec:future_selection_strategies_app}

This section gives an extended view of the four future condition strategies from Section~\ref{sec:future_selection_strategies}.

\paragraph{Static future conditioning}
Static future conditioning selects the first start timestep in the future based on the current timestep. 
Equation~\ref{eq:static_future} states the formula for selecting the starting timestep of the future trajectory. This type of future conditioning is not conditioned on the spatial information.

\paragraph{Closest future conditioning}

The closest future conditioning uses the agent's current position and finds the timestep of the closest point in the ground truth trajectory measured by Euclidean distance. The formula can be seen in Equation~\ref{eq:closest_future}. 

Closest future conditioning takes into account the spatial distance between the agent and the conditioned point. By finding the closest point to the agent, we are minimising the spatial distance, keeping the agent in the training distribution for as long as possible.

The downside of this method is the potential to get stuck in an infinite loop. By ignoring the temporal aspect of the trajectory following, the agent is unable to differentiate between two frames in the same spatial position. 

This can manifest in trajectories where the agent stands still for N frames and then starts moving afterwards. If N is larger than then the length of the trajectory used as an input to the agent, the spatial position of the goal state will be in the same spatial position as the starting state for which a trained IDM will do a no-op, since the goal is reached, causing an infinite loop, where the goal doesn't change, and the agent doesn't move, causing the goal not to change again.

\begin{figure}[!ht]
\centering
\includegraphics[width=.5\linewidth]{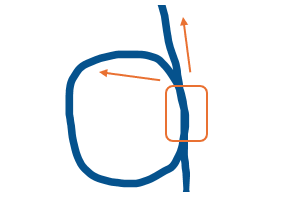}
\caption{Loop problem with closest future conditioning. Due to no temporal structure, if the trajectory contains two or more frames with the same location, the closest future conditioning strategy at the position outlined by the orange square doesn't know if the agent has already made the loop, and randomly chooses between the two possible future paths depicted by orange arrows.}
\label{fig:closest_strategy_loop_problem}
\end{figure}

Another example would be a trajectory containing loops, where the future selection strategy is not able to determine whether we are at the start or the end of the loop. Figure~\ref{fig:closest_strategy_loop_problem} demonstrates this situation. 

\paragraph{Radius future conditioning}
Radius future conditioning moves the future start timestep by 1 if the agent is sufficiently close, i.e. inside of the specified radius, to the current future start timestep, as defined in Equation~\ref{eq:radius_future}. 
Experimental results for different values for the radius are presented in Table~\ref{tab:radius_selection_ablation} in the Appendix.

Radius future conditioning takes into account both temporal and spatial distance. This approach successfully tackles the infinite loop examples in Figure~\ref{fig:closest_strategy_loop_problem} due to the constraint that the future start timestep can only be increased. 

\paragraph{Inner-Outer future conditioning}

The Inner-Outer future conditioning extends the Radius future conditioning by allowing it to progress the future for more than one frame. This method uses two radiuses, the inner radius and the outer radius. Experimental results for different values for the radius are presented in Table~\ref{tab:radius_selection_ablation}.

There are three cases in which the agent could be:
\begin{enumerate}
    \item Inside of the inner radius - We move the future until it leaves the inner radius. This causes the future conditioning to fix the issues where the agent can be ahead of the future conditioning due to incorrect actions, or FPS variation.
    \item Outside of the inner radius, but inside of the outer radius - We move the future by 1 frame, same as the Radius selection strategy.
    \item Outside of the outer radius - We do not move the future, same as the Radius selection strategy.
\end{enumerate}

\begin{table}[!ht]
    \caption{AUC results for different radius values}
    \label{tab:radius_selection_ablation}
    \centering
    \begin{tabular}{lccccc}
        \toprule
        & \multicolumn{2}{c}{Radius} & \multicolumn{3}{c}{Inner-Outer} \\
        \cmidrule(lr){2-3}\cmidrule(lr){4-6}
        Trajectory & R=800 & R=1600 & I=100,O=200 & I=200,O=400 & I=800,O=3200 \\
        \midrule
        Jumppad left    & \textbf{1.00} & \textbf{1.00} & 0.70 & 0.98 & \textbf{1.00} \\
        Jumppad right   & \textbf{0.97} & 0.94 & 0.38 & 0.98 & \textbf{1.00} \\
        Jumppad mid     & \textbf{1.00} & 0.97 & 0.72 & 0.98 & \textbf{1.00} \\
        Benchmark 0     & 0.28 & \textbf{0.48} & 0.05 & 0.10 &\textbf{ 0.79} \\
        Benchmark 1     & 0.99 & \textbf{1.00} & 0.19 & 0.08 & \textbf{1.00} \\
        Benchmark 2     & 0.69 & \textbf{0.94} & 0.13 & 0.16 & \textbf{0.79} \\
        Dojo Ramp      & 0.33 & \textbf{0.63} & 0.17 & 0.21 & \textbf{0.68} \\
        Dojo Gong       & 0.25 & \textbf{0.47} & 0.06 & 0.18 & \textbf{0.56} \\
        Mean            & 0.69 & \textbf{0.81} & 0.30 & 0.55 & \textbf{0.85} \\
        \bottomrule
    \end{tabular}
\end{table}

\subsection{Evaluation Metrics}\label{sec:auc}

This section defines the AUC metric used in the paper. For a reference trajectory $\tau$ and an agent rollout $\hat{\tau}$ represented by the state sequences $\tau = \{\bm{x}_t\}_{t=1}^T$ and $\hat{\tau}= \{\hat{\bm{x}}_t\}_{t=1}^T$ respectively, we define the coverage rate
\begin{equation}
    f(\hat{\tau}, \tau, r) =  \frac{1}{T}\sum_{t=1}^T\Big(\mathbb{I}(\| \bm{x}_t  - \hat{\bm{x}}_t\| < r)\Big),
\end{equation} 
as the percentage of timesteps where the agent rollout is closer to the reference trajectory than a specified radius $r$. The indicator function $\mathbb{I}(\| \bm{x}_t  - \hat{\bm{x}}_t\| < r)$ is 1 if $\| \bm{x}_t  - \hat{\bm{x}}_t\| < r$ and 0 otherwise.

The AUC metric is defined as the average coverage rate for radius values $r \in [0, R]$ below a maximum radius $R$. This average is proportional to the area under the curve of $f$, 
\begin{equation}
    AUC(\hat{\tau}, \tau) = \frac{1}{R}\int_0^R  f(\hat{\tau}, \tau, r)dr.
\end{equation}
where the maximum radius $R$ is defined as 
\begin{equation}
    R = max_{t \in [1, T]} \|\bm{x}_0 - \bm{x}_t\|. 
\end{equation}

On one hand, if the agent rollout is close to the reference trajectory, $f$ grows rapidly as $r$ increases resulting in a larger AUC value. On the other hand, if the agent rollout is not similar to the reference trajectory, for most values of $r$ the coverage rate remains low resulting in a lower AUC value.

Additionally, by rescaling using the maximum radius $R$, the AUC metric is less dependent on the spread of the reference trajectory. 

\subsection{Full results}
\label{sec:app:full-results}

This section shows the full results, per trajectory per model.

\begin{table}[!ht]
    \caption{FI results for the general setting evaluation on the 8 heldout trajectories. A higher FI is better.}
    \label{tab:full_f1_results_general}
    \centering
    \begin{tabular}{lcccccc}
        \toprule
        & \multicolumn{3}{c}{MLP} & \multicolumn{3}{c}{GPT} \\
        \cmidrule(lr){2-4}\cmidrule(lr){5-7}
        Trajectory & ConvNeXt & DINOv2 & WHAM & ConvNeXt & DINOv2 & WHAM \\
        \midrule
        Jumppad left    &  \textbf{1.00} & 0.89 & 0.94 & 0.96 & 0.98 & 0.88  \\
        Jumppad right   &  \textbf{1.00} & \textbf{1.00} & 0.87 & 0.97 & \textbf{1.00} & 0.88  \\
        Jumppad mid     &  \textbf{1.00} & 0.98 & 0.69 & \textbf{1.00} & 0.85 & 0.92  \\
        Benchmark 0     &  0.25 & 0.21 & 0.18 & \textbf{0.48} & 0.30 & 0.25  \\
        Benchmark 1     &  0.61 & 0.63 & 0.14 & \textbf{1.00} & 0.54 & 0.43  \\
        Benchmark 2     &  0.62 & 0.21 & 0.09 & \textbf{0.94} & 0.92 & 0.33  \\
        Dojo Ramp      &  0.22 & 0.21 & 0.22 & \textbf{0.29} & 0.24 & 0.22  \\
        Dojo Gong       &  0.09 & 0.11 & 0.12 & \textbf{0.18} & 0.12 & 0.09  \\
        Mean            &  0.55 & 0.53 & 0.40 & \textbf{0.73} & 0.62 & 0.50  \\
        \bottomrule
    \end{tabular}
\end{table}

\begin{table}[!ht]
    \caption{Ablation showing the performance of the ConvNeXt-GPT agent using different inputs to the model.}
    \label{tab:different-input-ablations-full}
    \centering
    \begin{tabular}{lcccccc}
        \toprule
          &  \multicolumn{2}{c}{Observations Only} & \multicolumn{2}{c}{Actions Only} & \multicolumn{2}{c}{Full Model} \\
        \cmidrule(lr){2-3}\cmidrule(lr){4-5}\cmidrule(lr){6-7}
        Trajectory & AUC  & FI   & AUC  & FI   & AUC  & FI   \\
        \midrule
        Jumppad left    & \textbf{0.99} & \textbf{1.00} & 0.93 & \textbf{1.00} & 0.99 & 0.96 \\
        Jumppad right   & \textbf{0.99} & \textbf{1.00} & 0.91 & 0.82 & 0.95 & 0.97 \\
        Jumppad mid     & \textbf{0.99} & 0.95 & 0.88 & 0.80 & 0.93 & \textbf{1.00} \\
        Benchmark 0     & 0.46 & 0.29 &	\textbf{0.53} & 0.17 & 0.50 & \textbf{0.48} \\
        Benchmark 1     & \textbf{0.98} & 0.92 &	0.65 & 0.41 & 0.81 & \textbf{1.00} \\
        Benchmark 2     & \textbf{0.93} & 0.88 &	0.45 & 0.25 & 0.69 & \textbf{0.94} \\
        Dojo Ramp      &\textbf{0.71} & 0.27 &	0.55 & 0.21 & 0.63 & \textbf{0.29} \\
        Dojo Gong       &\textbf{0.68} &\textbf{0.32} &	0.67 & 0.08 & 0.67 & 0.18 \\
        Mean & 0.84 & 0.70 & 0.69 & 0.47 & \textbf{0.86} & \textbf{0.73} \\
        \bottomrule
    \end{tabular}
\end{table}

\begin{table}[!ht]
    \caption{Ablation showing the performance of the ConvNeXt-GPT agent with varying future and past lengths.}
    \label{tab:future-past-length-performance-full}
    \centering
        \begin{tabular}{lcccccccccc}
        \toprule
              & \multicolumn{2}{c}{BC}   & \multicolumn{2}{c}{1P-1F}   & \multicolumn{2}{c}{10P-1F}   & \multicolumn{2}{c}{1P-10F} & \multicolumn{2}{c}{Full model}    \\
        \cmidrule(lr){2-3}\cmidrule(lr){4-5}\cmidrule(lr){6-7}\cmidrule(lr){8-9}\cmidrule(lr){10-11}
        Trajectory      & AUC  & FI   & AUC  & FI   & AUC  & FI  & AUC  & FI   & AUC  & FI   \\
        \midrule
        Jumppad left    & 0.90 & 0.78 &	0.99 & \textbf{1.00} & 0.99 & 0.96 & 0.99 & \textbf{1.00} & 0.99 & 0.96 \\
        Jumppad right   & 0.90 & 0.64 &	0.99 & \textbf{1.00} & 0.99 & 0.90 & 0.99 & \textbf{1.00} & 0.99 & 0.97 \\
        Jumppad mid     & 0.92 & \textbf{1.00} &	0.99 & 0.98 & 0.99 & 0.91 &	0.99 & 0.98 & 0.99 & \textbf{1.00} \\
        Benchmark 0     & 0.34 & 0.18 &	0.47 & 0.28 & 0.47 & 0.39 &	0.55 & 0.21 & 0.64 & \textbf{0.48} \\
        Benchmark 1     & 0.54 & 0.37 &	0.98 & 0.98 & 0.98 & \textbf{1.00} &	0.97 & \textbf{1.00} & 0.98 & \textbf{1.00} \\
        Benchmark 2     & 0.44 & 0.27 &	0.89 & 0.68 & 0.94 & 0.78 &	0.87 & 0.64 & 0.95 & \textbf{0.94} \\
        Dojo Ramp      & 0.55 & 0.21 &	0.72 & \textbf{0.45} & 0.73 & 0.40 & 0.54 & 0.23 & 0.65 & 0.29 \\
        Dojo Gong       & 0.51 & 0.11 &	0.69 & 0.18 & 0.72 & \textbf{0.21} &	0.44 & 0.04 & 0.70 & 0.18 \\
        Mean & 0.64 & 0.45 & 0.84 & 0.69 & 0.85 & 0.69 &	0.79 & 0.64 & 0.86 & 0.73 \\
        \bottomrule
    \end{tabular}
\end{table}

\end{document}